\documentclass[letterpaper, 10 pt, conference]{ieeeconf}
\IEEEoverridecommandlockouts  
\usepackage{bm}
\usepackage{tikz}
\usepackage{xargs}
\usepackage[hyphens]{url}
\usepackage{times}
\usepackage{array}
\usepackage{xcolor}
\usepackage{siunitx}
\usepackage{relsize}
\usepackage{amssymb}
\usepackage{amsmath}
\usepackage{flushend}
\usepackage{graphics}
\usepackage{graphicx}
\usepackage{hyperref}
\usepackage{multicol}
\usepackage{verbatim}
\usepackage{multirow}
\usepackage{booktabs}
\usepackage{pgfplots}
\usepackage{algorithm}
\usepackage{xinttools}
\usepackage{glossaries}
\usepackage{algorithmic}
\usepackage[nolist]{acronym} \newacro{vs}[VS]{Visual Servoing}
\newacro{pulp}[PULP]{Parallel Ultra-Low Power}
\newacro{dof}[DoF]{Degrees of Freedom}
\newacro{nn}[NN]{artificial Neural Network}
\newacro{cnn}[CNN]{Convolutional Neural Network}
\newacro{fcn}[FCN]{Fully Convolutional Network}
\newacro{fc}[FC]{Fully-Connected}
\newacro{ml}[ML]{Machine Learning}
\newacro{ssl}[SSL]{Self-Supervised Learning}
\newacro{ros}[ROS]{Robot Operating System}
\newacro{dl}[DL]{Deep Learning}
\newacro{mav}[MAV]{Micro Aerial Vehicle}
\newacro{uwb}[UWB]{Ultra-Wide Band}
\newacro{rss}[RSSI]{Received Signal Strength Intensity}
\newacro{ekf}[EKF]{Extended Kalman filter}
\newacro{auc}[AUC]{Area Under the Receiver Operating Characteristic Curve}
\newacro{soa}[SoA]{State-of-the-art}
\newacro{soc}[SoC]{System-on-Chip}

\newacro{ledp}[LED-P]{LED state prediction Pretext}
\newacro{decro}[EDNN]{Efficient Deep Neural Networks}
\newacro{bas}[BAS]{Baseline}
\newacro{ub}[UB]{Upper Bound}
\newacro{ae}[AE-P]{Autoencoding Pretext}
\newacro{clip}[CLIP]{Contrastive Language-Image Pre-training}

\usetikzlibrary{arrows.meta}
\usepgfplotslibrary{groupplots}
\usepgfplotslibrary{fillbetween}
\pgfplotsset{compat=1.4}
\graphicspath{{fig/}}
\setacronymstyle{long-short}
\DeclareMathAlphabet{\mathpzc}{OT1}{pzc}{m}{it}

\DeclareMathOperator{\thefunction}{\bm{wm_{\hat{P}}}}

\newcommand\copyrighttext{%
  \footnotesize \textcopyright 2024 IEEE. Personal use of this material is permitted.
  Permission from IEEE must be obtained for all other uses, in any current or future
  media, including reprinting/republishing this material for advertising or promotional
  purposes, creating new collective works, for resale or redistribution to servers or
  lists, or reuse of any copyrighted component of this work in other works.
  }
\newcommand\copyrightnotice{%
\begin{tikzpicture}[remember picture,overlay]
\node[anchor=south,yshift=10pt] at (current page.south) {\fbox{\parbox{\dimexpr\textwidth-\fboxsep-\fboxrule\relax}{\copyrighttext}}};
\end{tikzpicture}%
}

\title{\LARGE \bf
Learning to Estimate the Pose of a Peer Robot in a Camera Image\\by Predicting the States of its LEDs
}

\author{Nicholas Carlotti$^{1}$, Mirko Nava$^{1}$, and Alessandro Giusti$^{1}$%
\thanks{$^{1}$All authors are with the Dalle Molle Institute for Artificial Intelligence (IDSIA), USI-SUPSI, Lugano, 6962, Switzerland \texttt{nicholas.carlotti@idsia.ch}}%
\thanks{This work is supported by the Swiss National Science Foundation, grant number 213074.}%
}

\begin{document}

\maketitle
\copyrightnotice
\thispagestyle{empty}
\pagestyle{empty}


\begin{abstract}
We consider the problem of training a fully convolutional network to estimate the relative 6D pose of a robot given a camera image, when the robot is equipped with independent controllable LEDs placed in different parts of its body.  The training data is composed by few (or zero) images labeled with a ground truth relative pose and many images labeled only with the true state (\textsc{on} or \textsc{off}) of each of the peer LEDs.  The former data is expensive to acquire, requiring external infrastructure for tracking the two robots; the latter is cheap as it can be acquired by two unsupervised robots moving randomly and toggling their LEDs while sharing the true LED states via radio.
Training with the latter dataset on estimating the LEDs' state of the peer robot (\emph{pretext task}) promotes learning the relative localization task (\emph{end task}).
Experiments on real-world data acquired by two autonomous wheeled robots show that a model trained only on the pretext task successfully learns to localize a peer robot on the image plane; fine-tuning such model on the end task with few labeled images yields statistically significant improvements in 6D relative pose estimation with respect to baselines that do not use pretext-task pre-training, and alternative approaches. 
Estimating the state of multiple independent LEDs promotes learning to estimate relative heading.
The approach works even when a large fraction of training images do not include the peer robot and generalizes well to unseen environments.
\end{abstract}

\section*{Supplementary Material}
Videos, dataset and code of the proposed approach are available at \small{\url{https://github.com/idsia-robotics/ssl-pretext-multi-led}}.

\section{Introduction}\label{sec:intro}

Estimating the relative pose of a peer robot is a key capability for multi-robot systems~\cite{dorigo2021swarm}, with wide-ranging applications spanning surveillance~\cite{vrba2019onboard, mishra2020drone}, surveying or exploration~\cite{mcguire2019minimal, nguyen2019swarmathon}, acrobatic and light shows~\cite{dias2016onboard, vasarhelyi2018optimized, weng2023multi}.
Solving the problem using visual inputs typically involves training a convolutional neural network with images and associated ground truth labels, representing the true relative pose of the peer robot. Collecting such a dataset can be expensive: generating ground truth labels requires specialized tracking hardware, e.g., often a fixed infrastructure; at the same time, generalizing to different environments and conditions requires training with a dataset that is large and varied.
To minimize the need for expensive labeled training data for the pose estimation task, i.e., our \emph{end task}, one can train the model to solve an auxiliary \emph{pretext task}~\cite{jing2020self}.  A good pretext task has two distinctive characteristics: it promotes learning visual features that are relevant to the end task and uses ground truth labels that are easy to collect.
 
In previous work \cite{nava2024self}, we have shown that learning to estimate the shared boolean state of the LEDs on the peer robot (being either all turned on, or all off) is an effective pretext task to learn the end task of 2D localization of the peer robot in the image plane, also called detection in the literature.
This pretext task entails a powerful idea: predicting whether the peer robot's LEDs are on or off given an image requires understanding where the peer robot is in the image. 
The task can be trained using images labeled only with the true state of the peer robot's LEDs; this data is easy to get from a collaborative peer, e.g., via radio, and large datasets can be collected autonomously by the pair of robots in any environment, without the need for external infrastructure or supervision.

\begin{figure}[t]
    \centering
    \includegraphics[width=1.0\linewidth,trim={20mm 0 0 0},clip]{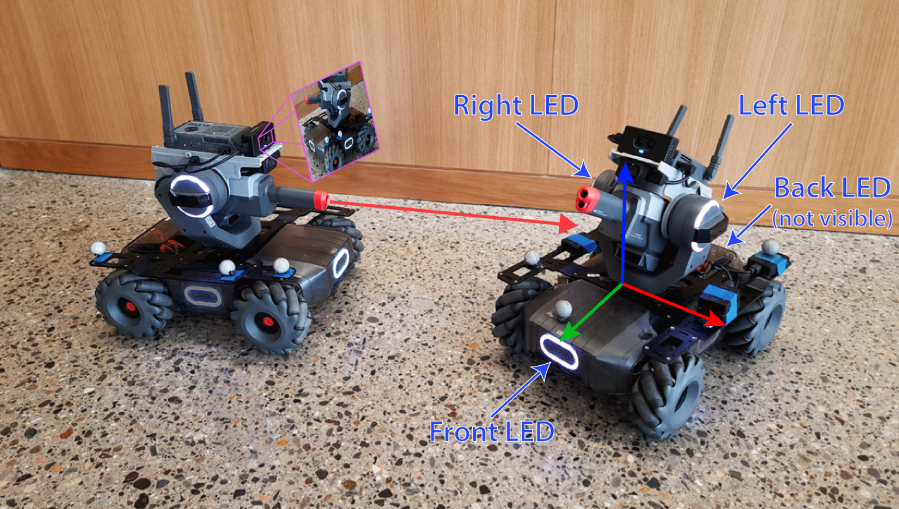}%
    \caption{A DJI RoboMaster S1 visually estimating the relative pose of a peer robot, using a model trained on an autonomously-collected dataset in which the four LED states of its peer robot are known for all images; instead, the true relative pose of its peer is known in few, or even zero images.}
    \label{fig:approach}
\end{figure}

This paper extends our previous work \cite{nava2024self} in multiple directions.
We show that the pretext task alone, \emph{without any labels for the end task}, leads to the model learning to localize the peer robot on the image plane. This is due to the inductive bias induced by our fully convolutional network~\cite{long2015fully} architecture.  It predicts the state of an LED as the weighted average of an output LED state map while the weights are given by an output robot position map. This weighting can be seen as an attention mechanism exploited for localization, similar to weakly supervised learning for object localization~\cite{zhangWeaklySupervisedObject2022a}, but using autonomously-collected LED state labels instead of semantic labels.
Instead of a single state shared by all of the peer's LEDs, our pretext task estimates \emph{four independent states}, one for each LED placed on a different side of the peer robot, as shown in Figure~\ref{fig:approach}.

\begin{figure}[h]
    \centering
    \includegraphics[width=1.0\linewidth]{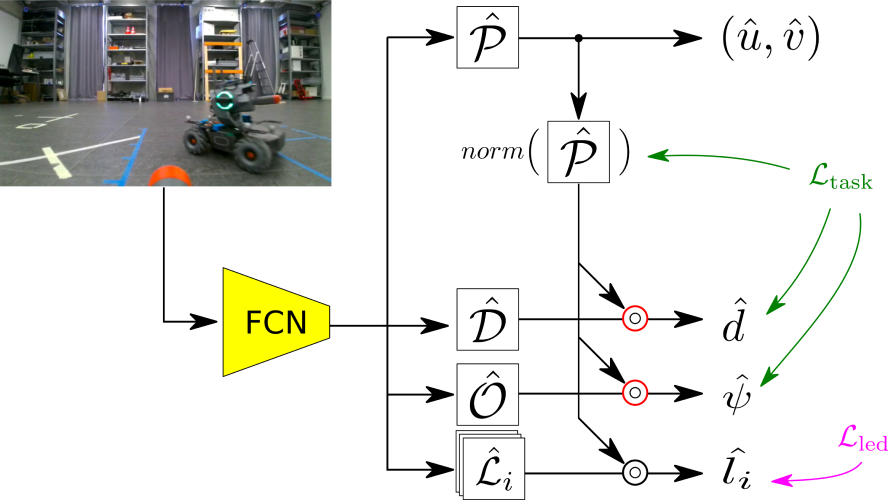}%
    \caption{Our fully convolutional network model takes an image as input and predicts maps for the robot's image space position $\bm{\hat{P}}$, distance $\bm{\hat{D}}$, orientation $\bm{\hat{O}}$ and state of the LEDs $\bm{\hat{L}}_i$. We obtain scalar values from the maps by element-wise multiplication (denoted as $\circ$) with $norm(\bm{\hat{P}})$, acting as an attention mechanism. By optimizing $\hat{l}_i$, the model learns to estimate the robot's LED state and position in the image; gradients for $\bm{\hat{P}}$ resulting from the optimization of $\hat{d}$ and $\hat{\psi}$ are blocked (see Section \ref{sec:model:losses} for details).}
    \label{fig:fcn}
\end{figure}

Solving the pretext task requires not only help to localize the peer in the image but also to understand its relative orientation, as shown in Figure~\ref{fig:fcn}.
Experiments confirm that the proposed pretext task helps to learn to estimate the peer's relative orientation.
Further, previous work~\cite{nava2024self} only considered the position of the peer robot in the image and assumed it to be always visible in images used to train the pretext task; here, we estimate the 6D relative pose of the peer and also consider the two robots autonomously collecting training data while roaming a confined environment: less than a quarter of the resulting images have the peer robot visible at all, with each LED being visible only for 8\% of images.
We show that the pretext task is effective in this challenging but realistic case.

Our \textbf{main} contribution is the improved methodology for an LED-based pretext task, presented in Section~\ref{sec:method}.  We extensively validate our approach on real-world data on the task of estimating the pose of a wheeled ground robot from the feed of an RGB monocular camera in Section~\ref{sec:setup}.
Results, reported in Section~\ref{sec:results}, show the ability of the robot to localize its peer without training on ground truth pose labels, and significant improvements when the model trained on the pretext task is fine-tuned for the end task on 10, 100 or 1000 labeled images; compared to a baseline that does not use the pretext task, and to a baseline trained on CLIP~\cite{radford2021learning} features, our approach significantly improves performance on 2D image localization, relative orientation estimation, and 3D pose estimation.  As a \textbf{secondary} contribution, we release the code and real-world dataset used for our experiments. Finally, we conclude the article and describe future work directions in Section~\ref{sec:conclusions}.

\section{Related Work}\label{sec:rw}

\subsection{Visual 6D Robot Pose Estimation}

Visual robot pose estimation problems are solved using hand-crafted~\cite{dias2016onboard, saska2017system, yuRobustRobotPose2019} or learned~\cite{leeCameratoRobotPoseEstimation2020, luPoseEstimationRobot2022} features.
Hand-crafted approaches propose algorithms to extract geometrical features of either paper-printed patterns \cite{saska2017system} or fiducial markers \cite{dias2016onboard} and process them to reconstruct the pose of the robot relative to the camera.
In the absence of markers or patterns, Yu et al.~\cite{yuRobustRobotPose2019} leverage dense information from an RGB-D camera to extract features and then localize the robot.
However, these techniques do not generalize to environmental settings not conceived at design time.
To address this limitation, practitioners resort to deep neural networks; these models are used for pose estimation of a robot arm's joints by predicting keypoints~\cite{leeCameratoRobotPoseEstimation2020}, and further optimize the keypoint placement~\cite{luPoseEstimationRobot2022}.

Our approach uses a self-supervised pretext task to train a deep neural network model, learning relevant visual features and enabling relative pose estimation of a peer robot given a single image.

\subsection{Self Supervised Learning for Visual Pose Estimation}

Training neural networks to solve complex visual tasks requires a large amount of labeled data.
Labeling this data is often expensive and time-consuming, involving specialized tracking hardware or tedious hand-labeling.
This issue is tackled by self-supervised learning techniques, where an autonomous robot collects its own training data~\cite{jang2018grasp2vec, deng2020self, li2022self} and uses it to train on a pretext task, learning features that are useful for solving perception problems~\cite{nava2024self, radosavovic2023real, nava2022learning}.
Automated data collection with a robot arm involves recording interactions with objects and the view of the scene, later used to learn object features~\cite{jang2018grasp2vec} or its pose, starting from models pre-trained in simulation and fine-tuned on real data~\cite{deng2020self}.
Nano-drone's poses are estimated from monocular cameras, training on labels generated from ultra-wide band sensors~\cite{li2022self}.

Approaches that rely on a pretext task in robotics use masked autoencoders to learn object pose estimation and manipulation~\cite{radosavovic2023real};
improve the detection of a drone by solving the pretext task of estimating the sound produced by its rotors~\cite{nava2022learning};
or the shared state of the drone's LEDs~\cite{nava2024self}, which are assumed to be either all on or off.
In contrast, the present work estimates the independent state of multiple robot-mounted LEDs: which of these are actually visible depends on the robot relative orientation.  This defines a more complex pretext task; learning to solve it leads to learning features that are informative not only for image-space localization, but also for estimating the relative orientation of the peer robot. 

Additionally, we demonstrate that our approach is robust to training with images that depict the robot infrequently, as it commonly occurs with a robot autonomously collecting data.

\subsection{Weakly Supervised Learning in Computer Vision}

In computer vision, weakly supervised learning focuses on reducing the need for lots of labeled examples for solving object localization~\cite{zhangWeaklySupervisedObject2022a} and segmentation~\cite{Zhang2019ASO} tasks.
The general framework consists in training classification models using coarse image labels, e.g., the textual description of the subject of a picture;
by training on these labels, models learn to extract finer information, such as image segmentations, by analyzing neural network activations \cite{wooCBAMConvolutionalBlock2018, dosovitskiy2021image} or employing feature clustering mechanisms \cite{pourianWeaklySupervisedGraph2015}.

The most notable approach that analyzes the network's hidden state for segmentation is Class Activation Mapping (CAM)~\cite{zhou2016learning}, and more recently Ablation-CAM~\cite{ramaswamy2020ablation}.
These approaches use a trained image classifier to detect discriminative areas of the input image, i.e., those responsible for making the model predict the image class.
For Ablation-CAM~\cite{ramaswamy2020ablation}, the activation maps are extracted with a gradient-free method, disabling feature maps and measuring its impact on the model's output.
Another proposed solution modifies a pre-trained classifier to predict a class for a set of regions partitioning the input image~\cite{bilen16Weakly}.
The individual scores for every region are then combined to allow for the detection of objects of interest in the input image.
Semantic segmentation is achieved by training a fully convolutional network to predict image labels using a weak supervision loss~\cite{pathak2015constrained}.
The loss enforces soft constraints based on the number of classes and number of elements within the class to improve segmentation results.
The introduction of attention mechanisms to visual models~\cite{wooCBAMConvolutionalBlock2018, dosovitskiy2021image} paved the way for novel weakly supervised segmentation approaches:
attention mechanism for convolutional neural networks have been proposed for segmenting images through the learning of classification tasks~\cite{maTriplestripAttentionMechanismbased2022,liangThreeDimensionAttentionMechanism2023};
while vision transformers segment images using a novel patch-based attention dropout layer~\cite{guptaViTOLVisionTransformer2022}.

The discussed approaches learn to detect objects in images by training on semantic class labels.
Here, instead, we employ the state of a robot's on-board LEDs as multi-class labels; this information can be easily collected by letting the robot broadcast the LEDs' state, enabling the automated collection of training data.
Further, our deep neural network model employs a soft attention mechanism: 
it designates one feature map as a mask applied to the other feature maps to predict the LEDs' state. After training, this map is used to locate the robot in the image.

\section{Method}\label{sec:method}

We focus on relative visual robot localization tasks where the goal is to estimate the relative pose of a target robot w.r.t. the frame of a peer robot, called observer, using as input the feed of a monocular camera.
To this end, we train a deep learning model using collected data, consisting of samples $\{\langle \bm{i}_j, \bm{p}_j, \bm{l}_j \rangle\}_{j=1}^{N}$ where $\bm{i} \in \mathbb{R}^{whc}$ is the image of $w \times h$ pixels and $c$ channels, $\bm{p} \in SE(3)$ is the 6D relative pose of the target robot consisting of $\langle x, y, z, \varphi, \theta, \psi \rangle$, and $\bm{l} \in \{\textsc{off}, \textsc{on}\}^{k}$ is the state of the robot's fitted LEDs consisting of $\langle l_1, \dots, l_k \rangle$.
Since our application involves a ground robot, $\phi$ and $\theta$ are assumed to be always zero.

We consider scenarios where the ground truth labels for the relative robot pose might only be measurable in dedicated lab environments with specialized hardware, e.g., tracking systems, while in other environments, this information in unknown and collected images might not feature a visible robot.
As such, the collected samples are said to be labeled if they feature known relative robot poses or unlabeled otherwise.
Based on this notion, we define three disjoint sets: the labeled training set $\mathcal{T}_\ell$ containing a limited amount of labeled samples, the unlabeled training set $\mathcal{T}_u$ containing unlabeled samples, and the testing set $\mathcal{Q}$ containing labeled samples.
For our experiments, we additionally consider the subset of the unlabeled testing set having samples with the robot visible, named $ \mathcal{T}_u^\nu$, and denote the full unlabeled set as $\mathcal{T}_u^a$. 

Using these samples, we learn a convolutional neural network $\bm{m_\theta}(\bm{i})$ parametrized by $\bm{\theta}$ and whose input is the image $\bm{i}$.
The model learns to predict the image-space position of the robot on the horizontal $u$ and vertical $v$ axes, also called 2D detection in the literature, the distance of the robot $d$ and its roll $\varphi$, pitch $\theta$, and yaw $\psi$ rotation.
Given the camera intrinsics, we reconstruct the 3D position $xyz$ of the robot by projecting a ray passing through the $uv$ pixel and selecting the point lying at distance $d$ from the camera.
As the architecture of choice, we propose a fully convolutional network (FCN) architecture~\cite{long2015fully}, being composed solely of convolution and pooling layers and producing maps of $w' \times h'$ cells as the output.
We take full advantage of the inductive bias of FCNs, namely that a cell's value depends only on a portion of the input image based on the model's receptive field, and that the weight-sharing of convolutions forces the model to recognize local patterns regardless of its location within the image.
Indeed, this helps the model to focus more on the robot's appearance and less on the background, and is a desirable property due to our expectation that the robot occupies a small portion of the field of view.

The model produces estimates for the position map $\bm{\hat{P}} \in [0, 1]^{w' h'}$, whose cells represent the likelihood of depicting a robot in the respective location;
the distance map $\bm{\hat{D}} \in [0, d_\text{max}]^{w'h'}$, representing the distance of the robot up to a maximum parametrized by $d_\text{max}$;
the orientation maps $\bm{\hat{O}} \in [-1, 1]^{w'h'2}$, representing the robot's orientation and consisting of sine $\bm{\hat{O}}_\text{sin}^\psi$ and cosine maps $\bm{\hat{O}}_\text{cos}^\psi
$;
and the LED state maps $\bm{\hat{L}} \in [0, 1]^{w'h'k}$, where the $i$-th map represents the state of $l_i$ and $1\!\le\!i\!\le\!k$.
We obtain scalar values from the model's predicted maps by weighting them with the predicted position map $\bm{\hat{P}}$ according to $\thefunction(\bm{X}) = \sum_{i=1}^{h'}\sum_{j=1}^{w'} \bigl( \text{norm}(\bm{\hat{P}})_{ij} \cdot \bm{X}_{ij} \bigr)$, where $\text{norm}(\cdot)$ normalizes the map s.t. its cells sum to one.
Specifically for orientation, we obtain the value of the angle from the predicted sine and cosine maps as $\hat{\psi} = \text{atan2} \bigl( \thefunction(\bm{\hat{O}}_\text{sin}^\psi), \thefunction(\bm{\hat{O}}_\text{cos}^\psi) \bigr)$.

\subsection{Losses}\label{sec:model:losses}
We train our model on one of two losses based on the availability of labels for the pose estimation task:
we define the downstream or end task loss as $\mathcal{L}_\text{task} = \frac{1}{3} (\mathcal{L}_\text{pos} + \mathcal{L}_\text{dist} + \mathcal{L}_\text{ori})$, trained on the labeled set $\mathcal{T}_\ell$, and the LED-based pretext loss $\mathcal{L}_\text{led}$, trained on $\mathcal{T}_u^a$ or the visible subset $\mathcal{T}_u^\nu$; all losses are designed to be bounded between zero and one.
To compute the position loss we first generate the ground truth map $\bm{{P}}$. To do this, we set cells in a circle of radius $r$ centered in $uv$ to one while the remaining cells are set to zero. We then calculate the element-wise product between the normalized position map and the ground truth one using $\thefunction$. By doing this, we measure the portion of pixels with high value in the predicted map $\bm{\hat{P}}$ that share the same coordinates with high-valued elements in $\bm{P}$.
This loss reaches zero only when predicted and ground truth maps perfectly overlap.
\begin{equation}\label{eq:lpos}
\mathcal{L}_\text{pos} = 1 - \thefunction(\bm{P})
\end{equation}
The distance loss uses $\thefunction$ to recover $\hat{d}$ and then computes the mean squared error with the ground truth distance
\begin{equation}\label{eq:ldist}
\mathcal{L}_\text{dist} = \frac{1}{d_\text{max}^2}\text{mse}\bigl(d, \thefunction(\bm{\hat{D}})\bigr)
\end{equation}
The orientation loss is defined as the sum of the mean squared error on the sine and cosine maps of each angle
\begin{equation}\label{eq:lori}
\mathcal{L}_\text{ori} = \frac{1}{4}
\sum_{f \in \{ \text{sin}, \text{cos} \} } \text{mse}\bigl( f(\psi), \thefunction(\bm{\hat{O}}_f^\psi) \bigr)
\end{equation}
The LED state loss is defined as the mean of the binary cross entropy for each of the $k$ robot LEDs
\begin{equation}\label{eq:lled}
\mathcal{L}_\text{led} = \frac{1}{k} \sum_{i=1}^{k} \text{bce} \bigl( l_i, \thefunction(\bm{\hat{L}}_i) \bigr)
\end{equation}

It is important to note that in $\mathcal{L}_\text{led}$, we allow gradients to flow through the predicted position map $\bm{\hat{P}}$, enabling the optimization of the robot position map without using labels;
this results in an attention-like mechanism, where the position map weights the cells of the LED map, assigning higher values to pixels containing relevant information for estimating the LEDs' state, i.e., those depicting the robot.
Instead, $\mathcal{L}_\text{dist}$ and $\mathcal{L}_\text{ori}$ block the gradients from flowing through the predicted position map $\bm{\hat{P}}$;
this avoids having the gradients for the $\bm{\hat{P}}$ map applied twice when optimizing $\mathcal{L}_\text{task}$.

\section{Experimental Setup}\label{sec:setup}
The proposed approach is applied to the problem of peer-to-peer localization of DJI S1 RoboMasters, omnidirectional ground robots equipped with four Swedish wheels and a front-facing monocular camera with a resolution of $640 \times 360$ pixels mounted on a pan-and-tilt turret.
The robot features six RGB LEDs: four are mounted on the base of the robot, one for each cardinal direction, and two are mounted on the left and right sides of the robot's turret; in our experiments we consider four LEDs: the two turret ones and the front and back ones of the robot base, while the left and right base LEDs are always off and ignored in our evaluation.

\begin{figure}[htb]
    \centering
    \includegraphics[width=1.0\linewidth]{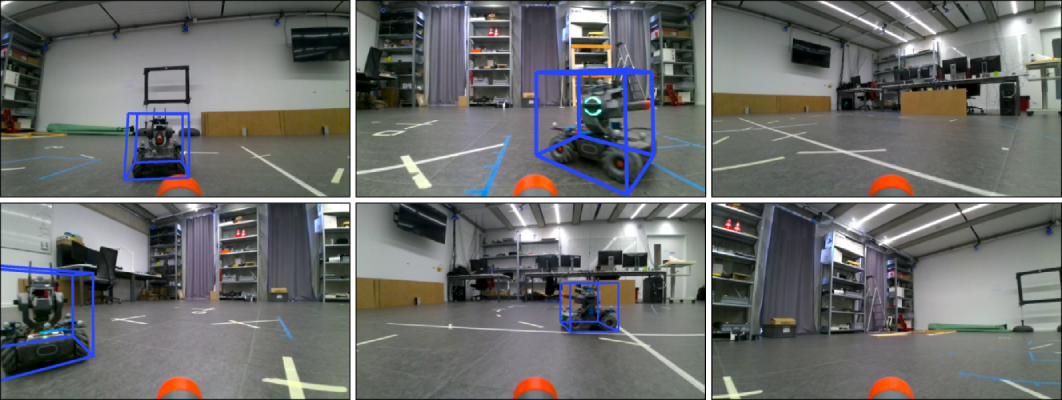}%
    \caption{Six samples from the unlabeled training set $\mathcal{T}_u^a$, where only 22\% of images feature a visible robot; ground truth robot poses are depicted with a blue bounding box.}
    \label{fig:dataset}
\end{figure}

\subsection{Dataset}\label{sec:setup:datasets}

The data collection setup consists of two robots roaming a $4 \times 4$ meters area inside our lab with a random policy: it keeps robots moving in a straight line until they collide with the edges of the area or with one another;
when it happens, they rotate in-place in one of the two directions by \ang{90} $\pm\ \mathcal{U}($\ang{-20}, \ang{20}$)$.
Every five seconds, each of the four LEDs of the target robot is randomly toggled, and its state is stored.
During data collection, we keep the turret at a fixed relative orientation w.r.t. the base of the robot, gathering images at 3 fps and the precise pose of the robots using a motion-tracking system featuring 18 infra-red cameras.

In total, we collected 37k samples over two recording sessions lasting approximately 1 hour each.
The samples are divided into 34k for the unlabeled training set $\mathcal{T}_u^a$, out of which only 7k have the robot visible (22\%) in them, 1k for the labeled training set $\mathcal{T}_\ell$, and 2k for the testing set $\mathcal{Q}$, the last two featuring the robot always visible;
we show samples from $\mathcal{T}_u^a$ in Figure~\ref{fig:dataset}.

\subsection{Training Strategies}\label{sec:setup:strategies}

We devise different training strategies based on the availability of robot pose labels.
With no access to labels, we train the model with $\mathcal{T}_u^\nu$ on the LED state estimation pretext task with $\mathcal{L}_\text{led}$, learning features that are also useful for pose estimation; additionally, we explore the effectiveness of the LED-based pretext task by training with the entire unlabeled training set $\mathcal{T}_u^a$, including the predominant type of image featuring no robot visible (88\%).

We compare our pretext strategy with Ablation-CAM,  a gradient-free CAM technique producing a score for each pixel of the input image.
High scores are assigned to areas of the image responsible for making the model predict the image class.
In weakly supervised learning, this technique is used to localize objects from a pre-trained classification model~\cite{zhangWeaklySupervisedObject2022a}.
This strategy involves training a custom MobileNet-V2 model~\cite{sandler2018mobilenet}, featuring 380k parameters, on the LED state estimation pretext task with $\mathcal{T}_u^\nu$.
Then, the model is fed to Ablation-CAM to generate the robot image-space position based on the four class activation maps (one for each LED).
Given these maps, we obtain the robot's position in the image as the average coordinates of the pixel with maximum value within each map.

Assuming access to a limited amount of labeled data, the most straight-forward strategy is to directly train with $\mathcal{T}_\ell$ on the pose task loss $\mathcal{L}_\text{task}$, representing a baseline.
We compare this baseline with the strategy that takes the pretrained model on the LED state estimation pretext task and transfers its skills to the downstream or end task, training with $\mathcal{T}_\ell$ on the pose task loss $\mathcal{L}_\text{task}$.
Additionally, we compare it against a different pretext approach of Contrastive Language-Image Pre-training (CLIP).
CLIP is a vision-language model trained on a large corpus of data to learn a similar encoding for images and associated captions, whose performance has been shown to outperform supervised counterparts on multiple tasks~\cite{radford2021learning}.
Specifically, we consider features extracted by the image encoder, and train a small feed forward neural network with $\mathcal{T}_\ell$ on the pose task loss $\mathcal{L}_\text{task}$ to predict the robot pose from CLIP's features.

Finally, we consider an upperbound model trained with $\mathcal{T}_u^* \cup \mathcal{T}_\ell$ on the pose task loss $\mathcal{L}_\text{task}$, representing the best performance achievable with our setup under the assumption of having unlimited access to robot pose labels, even for $\mathcal{T}_u^*$ marked with an asterisk to denote the inclusion of pose labels.

\subsection{Neural Network Training}\label{sec:setup:training}

In our experiments, we consider the pose of a ground robot; as such, we only consider the model performance on the heading $\psi$ and modify $\mathcal{L}_\text{ori}$ accordingly.
Additionally, we estimate the state of $k = 4$ LEDs and set the maximum distance parameter $d_\text{max} = 5$ meters, as robots do not exceed this relative distance in the collected data.
All strategies train on their respective datasets for 100 epochs with the Adam optimizer, using a cosine annealing learning rate schedule starting at $10^{-3}$ and decaying to $10^{-5}$ at the end of the training, and a batch size of 64 samples.
We employ the following data augmentations to increase the variability of the images: additive simplex noise; random translation and rotation ($\pm$ 64 px, $\pm$ \ang{9}); and random hue shift, brightness and contrast.
FCN models produce output maps of size $w' \times h' = 80 \times 45$ pixels;
they are evaluated using weights $\bm{\theta}$ at the last epoch of training with the exception of upperbound, which reaches its best performance after 40 epochs.

\subsection{Evaluation Metrics}\label{sec:setup:metrics}

Models trained using the different strategies are evaluated on the testing set $\mathcal{Q}$.
For the LED state estimation task, we consider the area under the receiver-operator characteristic curve (AUC) averaged over the four LEDs.
An ideal model has an AUC of 100\%, while a naive classifier scores 50\%.
We introduce an evaluation metric for each of the pose task losses: for the 2D detection task, we report the median euclidean distance between ground truth $uv$ and predicted $\widehat{uv}$ pixel position of the robot in the image plane $E_{uv}$;
for the distance task, we report the median absolute difference of ground truth $d$ and predicted $\hat{d}$ relative robot distance $E_{d}$;
and for the orientation task, we report the median circular distance between ground truth $\psi$ and predicted $\hat{\psi}$ robot heading $E_{\psi}$.
The circular distance measures the difference between two angles accounting for the discontinuity at zero.
Additionally, we consider the median of the euclidean distance of the ground truth $xyz$ and predicted $\widehat{xyz}$ 3D position of the robot $E_{xyz}$, computed from $\widehat{uv}$ and $\hat{d}$ using the camera intrinsics, as described in Section~\ref{sec:method}.

\section{Results}\label{sec:results}

\begin{table*}[th]
    \setlength\tabcolsep{1.2mm} 
    \renewcommand{\arraystretch}{1.2} 
    \centering
    \caption{Model's median error on 2D position $uv$, heading $\psi$, distance $d$, and 3D position $xyz$ averaged on four replicas per row.}
    \label{tab:table}
    \begin{tabular}{lccrrrr>{\centering\arraybackslash}p{60mm}}
    \toprule
    \multirow{2}{*}{Model} & \multicolumn{2}{c}{Training set for task} & \multicolumn{1}{c}{$E_{uv}$} & \multicolumn{1}{c}{$E_{\psi}$} & \multicolumn{1}{c}{$E_{d}$} & \multicolumn{1}{c}{$E_{xyz}$} & {Point plot for $E_{uv}$ [px] $\leftarrow$} \\
    & Pretext & Downstream & [px] $\downarrow$ & [deg] $\downarrow$ & [cm] $\downarrow$ & [cm] $\downarrow$ & Error bars denote 95\% conf. int. \\
    \midrule
    Dummy Mean       &                 $-$ &                           $-$ & 122 & 80.1 &  67.2 & 103.0 & \multirow{1}{*}{\input{tableplot.tikz}} \\
    Pretext          & $\mathcal{T}_u^\nu$ &                           $-$ &  51 &  $-$ &   $-$ &   $-$ & \\
    Ablation-CAM~\cite{ramaswamy2020ablation} &  $\mathcal{T}_u^\nu$ & $-$ &  33 &  $-$ &   $-$ &   $-$ & \\
    Upperbound       & $-$ &       $\mathcal{T}_u^* \cup \mathcal{T}_\ell$ &  11 & 22.2 &  18.7 &  31.3 & \\
    \cmidrule{1-7}
    Downstream-1000  & $\mathcal{T}_u^\nu$ &     $\mathcal{T}_\ell^{1000}$ &  15 & 40.3 &  22.2 &  40.0 & \\
    Baseline-1000    &                 $-$ &     $\mathcal{T}_\ell^{1000}$ &  26 & 70.1 &  36.1 &  59.3 & \\
    CLIP-1000~\cite{radford2021learning} & $-$ & $\mathcal{T}_\ell^{1000}$ & 133 & 94.8 & 129.7 & 149.3 & \\
    Downstream-100   & $\mathcal{T}_u^\nu$ &     ~$\mathcal{T}_\ell^{100}$ &  25 & 76.5 &  41.5 &  61.8 & \\
    Baseline-100     &                 $-$ &     ~$\mathcal{T}_\ell^{100}$ & 149 & 93.2 &  67.0 & 136.3 & \\
    Downstream-10    & $\mathcal{T}_u^\nu$ &     ~~$\mathcal{T}_\ell^{10}$ &  38 & 93.2 &  57.7 &  92.0 & \\
    Baseline-10      &                 $-$ &     ~~$\mathcal{T}_\ell^{10}$ & 190 & 95.8 &  66.2 & 152.3 & \\
    \cmidrule{1-7}
    Downstream-$a$-1000 & $\mathcal{T}_u^a$ &    $\mathcal{T}_\ell^{1000}$ &  14 & 64.1 &  38.1 &  66.3 & \\
    Downstream-$a$-100  & $\mathcal{T}_u^a$ &    ~$\mathcal{T}_\ell^{100}$ &  27 & 84.0 &  62.5 &  85.0 & \\
    Downstream-$a$-10   & $\mathcal{T}_u^a$ &    ~~$\mathcal{T}_\ell^{10}$ &  76 & 88.4 &  79.7 & 121.8 & \\
    Pretext-$a$         & $\mathcal{T}_u^a$ &                          $-$ &  45 &  $-$ &   $-$ &   $-$ & \\[1.5mm]
    \bottomrule
    \end{tabular}
\end{table*}

In this section, we report a detailed analysis of our experimental results.
Specifically, we show that the proposed approach enables learning of the robot detection task without position labels in Section~\ref{sec:results:unlabeled}.
The trained models are then transferred to the 3D position and heading tasks, investigating how the amount of labeled training data affects performance in Section~\ref{sec:results:downstream}, and the role of the pretext in learning the robot's heading in Section~\ref{sec:results:oreintation}.

The robustness of the pretext approach is explored in Section~\ref{sec:results:invisible} by training on the unlabeled dataset $\mathcal{T}_u^a$ in which only 22\% of the images depict the target robot, whereas the rest only feature the background.
Finally, we test the generalization ability by deploying the model in never-seen-before environments in Section~\ref{sec:results:generalization}.

\subsection{LED Pretext Task Learns Robot Detection Without Labels}\label{sec:results:unlabeled}

We report in the first panel of Table~\ref{tab:table} the performance of our pretext strategy compared with a dummy mean and the upperbound.
Without having access to robot pose labels, the pretext model has learned to detect the robot in the image-plane simply by optimizing the LED-based pretext loss $\mathcal{L}_\text{led}$.
Remarkably, it achieves a median 2D position error of only 51 pixels for an input image of $640 \times 480$ pixels, a considerable feat given the very limited supervision it received.
This result indicates that our strategy of weighting the LED maps with the position map, combined with the inductive bias of FCNs, described in Section~\ref{sec:method}, induces the model to recognize the area of the image where information of the LEDs' state is found, i.e., the area where the robot is depicted.

Comparing pretext with Ablation-CAM, we note that our approach is outperformed by Ablation-CAM on robot detection;
however, this strategy does not lend itself to skill transfer to a downstream task as it requires pre-training a separate classification model.
In addition, this strategy is not feasible for real-time inference: on a NVIDIA RTX 2080, Ablation-CAM requires about 10 seconds per image, while our FCN model takes only 2.1 ms.

\subsection{LED Pretext Task Transfers Well to Pose Estimation}\label{sec:results:downstream}

In the second panel of Table~\ref{tab:table}, we compare the performance of the baseline strategy with the downstream one, the only difference being the initial weights: for baseline they are randomly initialized, while for downstream we use the weights of the pretext model.
Additionally, we investigate how the size of the labeled training set affects the performance by training the two strategies with 10, 100 or 1000 samples from $\mathcal{T}_\ell$; we represent these sets respectively as $\mathcal{T}_\ell^{10}$, $\mathcal{T}_\ell^{100}$, and $\mathcal{T}_\ell^{1000}$.
Regardless of the amount of labels, the downstream model outperforms the baseline counterpart on all metrics;
in particular, the two score closest when training on $\mathcal{T}_\ell^{1000}$, where the downstream model achieves a statistically significant%
\footnote{P-values computed using the one tailed Welch's T-test on four replicas.}
improvement on the median $uv$ error of 11 pixels, p-value=0.005, and on $\psi$ by \ang{29.8}, p-value$<$0.001.

More interestingly, the downstream model trained with $\mathcal{T}_\ell^{100}$ achieves the same performance on $uv$ of the baseline model trained with $10\times$ the amount of labeled data, and the downstream model trained with $\mathcal{T}_\ell^{1000}$ reduces the gap to just 4 pixels of $uv$ error with the upperbound model, which is trained on 35k labeled samples.
Baseline models trained on $\mathcal{T}_\ell^{10}$ and $\mathcal{T}_\ell^{100}$ perform equal or worse to the dummy mean model, while the downstream model trained on $\mathcal{T}_\ell^{100}$ outperforms dummy on all metrics and the downstream model trained on $\mathcal{T}_\ell^{10}$ on all metrics except for $\psi$.
Comparing with CLIP demonstrates that our LED pretext better captures geometrical information about the robot;
instead, CLIP's language-image pretext does not lend itself well to pose estimation, and further lacks the inductive bias that our FCN architecture posses, resulting in a worse performance than baseline trained on the same data.

On the 3D position, the downstream model consistently outperforms the respective baseline model across all percentages of $\mathcal{T}_\ell$ in a statistically significant manner, with the closest example being of models trained on $\mathcal{T}_\ell^{1000}$ where p-value=0.013.

\begin{figure}[tbh]
    \centering
    \includegraphics[width=1.0\linewidth]{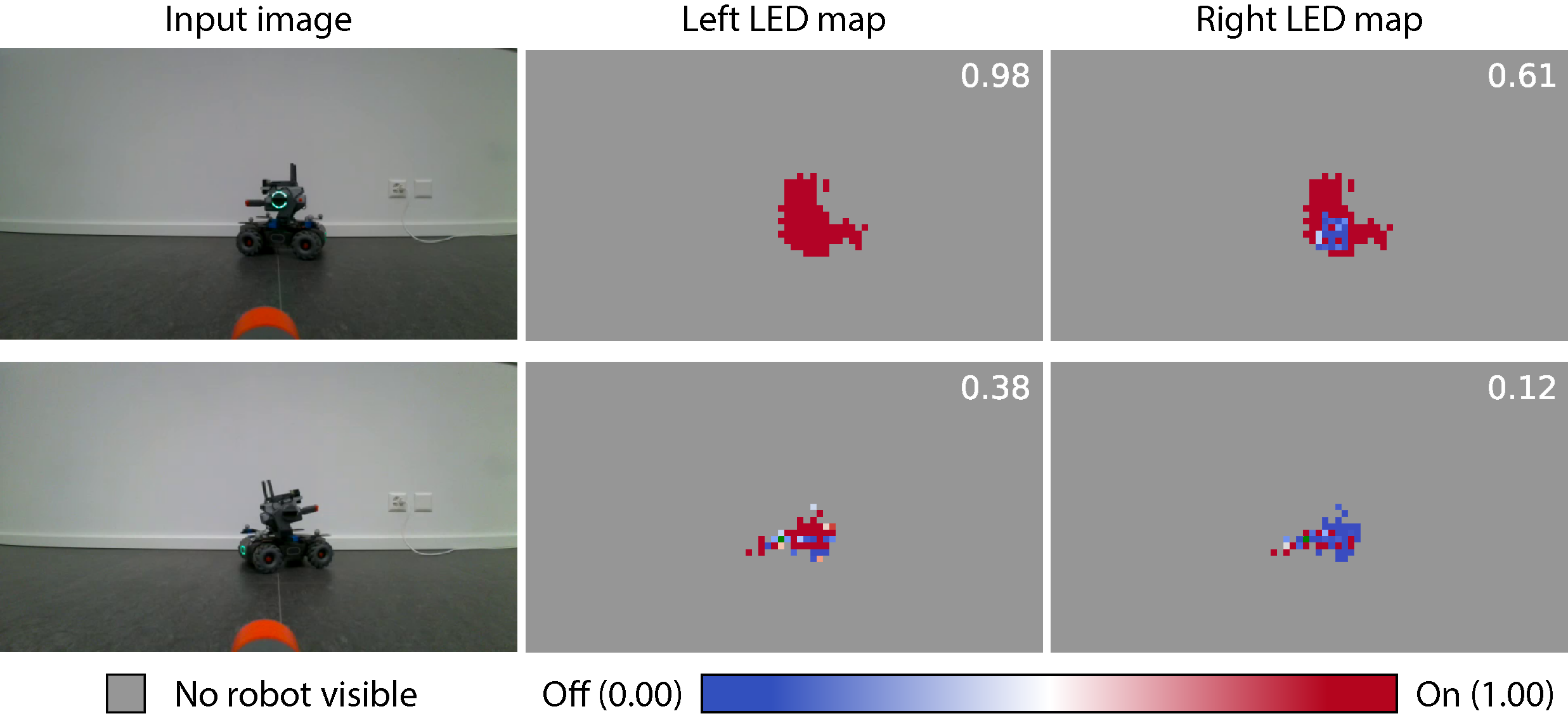}%
    \caption{Predicted LED maps by the pretext model trained on random 3.5k samples from $\mathcal{T}_u^\nu$. From left to right, input image, left LED map, and right LED map. Top image depicts the robot with left LED visible and turned on; bottom image depicts right LED visible and turned off. Position map cells with low probability of having a robot are depicted in gray. Predicted LED state scalars are reported on the top right corners of the LED maps.}
    \label{fig:confidence}
\end{figure}

\begin{figure*}[bht]
    \centering
    \includegraphics[width=1.0\linewidth]{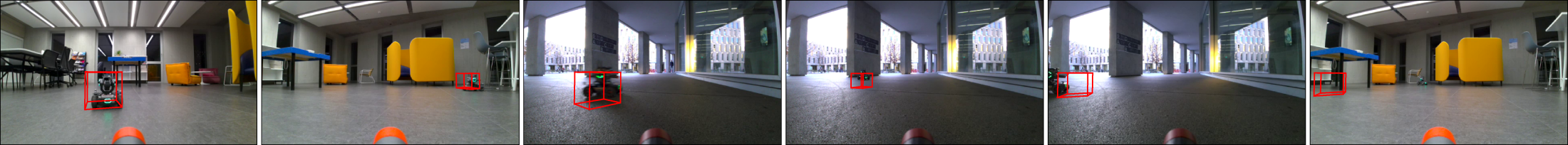}%
    \caption{Six images from unseen environments and predicted robot's bounding boxes (red) generated by the downstream model trained on $\mathcal{T}_\ell^{1000}$.}
    \label{fig:generalization}
\end{figure*}

\subsection{LED Pretext Task Captures Robot Heading}\label{sec:results:oreintation}

So far, we considered each LED to be independent from the others and speculated that this helps the pretext in learning useful features for the robot's heading.
In the following, we address this point with an experiment in which we consider the downstream training strategy using a subset of samples of $\mathcal{T}_u^\nu$.
We compare two models pretrained on the pretext task: the first uses 3.5k samples where the left and right turret LEDs have the same state, while the second uses 3.5k random samples from the same dataset;
the models are then transferred to the downstream task of pose estimation using $\mathcal{T}_\ell^{1000}$.
We report a median heading estimation error of \ang{45.9} for the downstream model trained with synchronized LEDs, and an error of \ang{35.7} for the downstream model trained with all possible LEDs combination.
Using the same statistical test as in Section~\ref{sec:results:downstream}, we report a statistically significant improvement on $\psi$ w.r.t. the synchronized LEDs pretext training, with a p-value of 0.003 obtained from four replicas of the experiment.

To further highlight the understating of the robot's heading achieved with the LED-based pretext training, we show in Figure~\ref{fig:confidence} the left and right LED maps when the pretext model is fed images with the robot placed on its side.
The maps for the visible LED are very consistent in its values, showing high confidence, while the maps for the invisible LED are divergent, showing uncertainty about the state; this indicates that our strategy does allow a model to capture the robot's orientation.

\subsection{LED Pretext Task is Robust to Images Without Robots}\label{sec:results:invisible}

In the last panel of Table~\ref{tab:table}, we report the performance of our approach when considering the entirety of $\mathcal{T}_u^a$, which features the robot visible only for 22\% of images, based on trajectories of the random controller described in Section~\ref{sec:setup:datasets}.
The median $uv$ error of the pretext and downstream models are in line with their counterparts trained using only images with visible robot of $\mathcal{T}_u^\nu$, highlighting the strength of this simple approach when applied to collected data without any filtering.
The robot heading and distance estimation suffer from training with a large quantity of images with no robot visible, degrading the performance of downstream models.

The notable exception is the downstream-$a$-10 model, which shows large instability during training and has a higher error than the downstream-10 model, at 76 pixels.
The cause is the different $\bm{\hat{P}}$ map's behavior in the pretext and downstream models:
in the former case, the $\bm{\hat{P}}$ map activates over a large area around the center of the robot; in the latter, the activations are more punctual and centered on the robot's body.
By fine-tuning the model on the downstream task, the position map $\bm{\hat{P}}$ transitions from a coarser and broader response to a more punctual one.
Transferring the model's skills from the pretext to the downstream pose task using only the 10 samples contained in $\mathcal{T}_\ell^{10}$ is not enough to correctly drive the optimization, whereas more samples are enough.

\subsection{LED Pretext Task Generalizes to Unseen Environments}\label{sec:results:generalization}

To test the generalization ability of our approach and the effect of the LEDs' state on pose estimation, we show in Figure~\ref{fig:generalization} the prediction of the downstream model trained with $\mathcal{T}_\ell^{1000}$ on data collected in never-seen-before environments.
Despite the different visual appearance of environments, the model correctly predicts the robot pose. 
The environments highlight the increase in difficulty when estimating the robot's distance and heading, which show larger errors than image-space detection.
Failure cases occur when the robot is further away than the maximum of 5 meters of the training set and when it is partially visible in the field of view.

\section{Conclusions}\label{sec:conclusions}

We presented a self-supervised learning approach for peer-to-peer robot pose estimation, leveraging robot LEDs to define a pretext task in order to learn useful features for pose estimation.
By predicting the state of LEDs from a given image, our approach improves in the estimation ability when compared to a baseline trained using the same amount of labeled data.

Results show that the approach is able to capture the 3D position and heading of ground robots, even when trained on images that feature the robot visible only 22\% of the times.
It also shows good generalization ability to never-seen-before environments.

Interestingly, our FCN model trained on the LED pretext using the attention-like mechanism has led the model to learning robot localization in images without using ground truth pose labels with a fair accuracy.
Future work will focus on exploring pose estimation without ground truth labels, improving the presented results in terms of performance, and integrating additional supervision sources.


\bstctlcite{IEEEexample:BSTcontrol}

\bibliographystyle{IEEEtran}
\bibliography{references}

\end{document}